\documentclass[10pt,twocolumn,letterpaper]{article}

\usepackage{iccv}
\usepackage{times}
\usepackage{epsfig}
\usepackage{graphicx}
\usepackage{amsmath}
\usepackage{amssymb}

\makeatletter
\@namedef{ver@everyshi.sty}{}
\makeatother
\usepackage{tikz}
\usetikzlibrary{positioning}
\usepackage{pgfplots}
\usepackage{pgfplotstable}
\usepackage{comment}
\usepackage{color}
\usepackage{booktabs}
\usepgfplotslibrary{fillbetween}
\definecolor{c_b0}{HTML}{A4DEF9}
\definecolor{c_b0.1}{HTML}{D2CCA1}
\definecolor{c_b0.5}{HTML}{387780}
\definecolor{c_b1}{HTML}{E83151}
\definecolor{c_b5}{HTML}{CFBAE1}
\definecolor{c_rnn}{HTML}{313715}
\definecolor{c_vpbs}{HTML}{E83151}
\definecolor{c_vgbs}{HTML}{0096C7}
\definecolor{c_tracktor}{HTML}{000000}
\definecolor{c_centertrack}{HTML}{000000}
\definecolor{c_kalman}{RGB}{0,50,255}
\definecolor{c_best_mean}{RGB}{255,125,0}

\usepackage[pagebackref=true,breaklinks=true,letterpaper=true,colorlinks,bookmarks=false]{hyperref}

\iccvfinalcopy 

\ificcvfinal\fi

\begin{document}

\title{Multiple Object Tracking with Mixture Density Networks for Trajectory Estimation \vspace{-1mm}}

\author{Andreu Girbau$^{1,2}$, Xavier Giró-i-Nieto$^1$, Ignasi Rius$^2$ and Ferran Marqués$^1$
\vspace{2mm}\\
$^1$Universitat Polit\`ecnica de Catalunya \:\:\:
$^2$AutomaticTV 
\\
\small{\{andreu.girbau, xavier.giro, ferran.marques\}@upc.edu, irius@mediapro.tv}}

\maketitle
\begin{abstract}

Multiple object tracking faces several challenges that may be alleviated with trajectory information.
Knowing the posterior locations of an object helps disambiguating and solving situations such as occlusions, re-identification, and identity switching.
In this work, we show that trajectory estimation can become a key factor for tracking, and present \textit{TrajE}, a trajectory estimator based on recurrent mixture density networks, as a generic module that can be added to existing object trackers. 
To provide several trajectory hypotheses, our method uses beam search. Also, relying on the same estimated trajectory, we propose to reconstruct a track after an occlusion occurs. 
We integrate TrajE into two state of the art tracking algorithms, CenterTrack \cite{zhou2020tracking} and Tracktor \cite{bergmann2019tracking}. Their respective performances in the MOTChallenge 2017 test set are boosted 6.3 and 0.3 points in MOTA score, and 1.8 and 3.1 in IDF1, setting a new state of the art for the CenterTrack+TrajE configuration.


\vspace{-5mm}
\end{abstract}
\section{Introduction}
\label{sec:intro}

The Multiple Object Tracking (MOT) task aims to estimate tracks of multiple objects across a sequence of frames. 
These objects must be detected with an accurate bounding box and maintain their identity over time. Tracking all the objects over a sequence can be useful in many applications, such as autonomous driving, robotics, and sports analytics.



In object tracking, predicting the position of an object can assist in many sub-tasks, such as re-identification or track association. 
Motion estimators are popular among the MOT community.
Some examples are the Kalman filter, optical flow, or constant velocity assumption, which are usually implemented to forecast the displacement of an object over consecutive frames.
We believe that modeling the objects trajectory, in contrast with the displacement between frames, can become a key factor of a tracker's performance.



We propose \textit{TrajE}, a trajectory estimator based on a recurrent mixture density network, that learns to estimate the underlying distribution of an object trajectory. By sampling from such a distribution, multiple hypotheses for the most likely position of the object can be forecasted, giving the tracker a prior on the subsequent object position. We combine this model with a beam search technique in order to explore multiple trajectory hypotheses per object. We provide a lightweight implementation of the model, designing TrajE as a single layer recurrent neural network, capable of estimating the objects trajectories.

\begin{figure}[t]
    \centering
    \includegraphics[width=\linewidth]{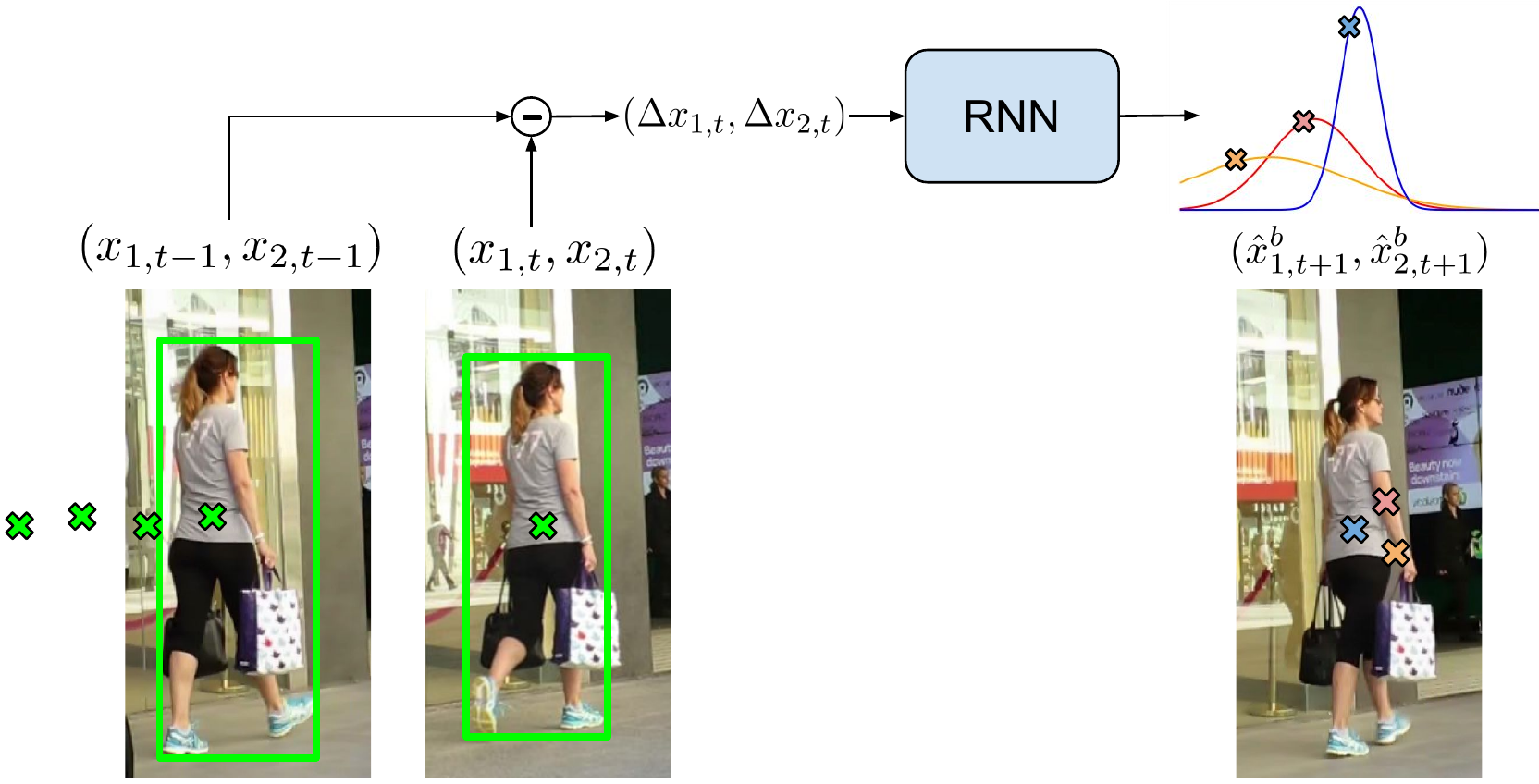}
    \caption{Concept of TrajE: our model to estimate trajectories. We train a Recurrent Neural Network to estimate a mixture of Gaussians that model the object trajectory. From it, several hypotheses (B) of the object position in the next time step are sampled.}
    \label{fig:fig_intro}
    \vspace{-3mm}
\end{figure}

We have trained \textit{TrajE} with pedestrian trajectories in the MOT challenge dataset \cite{MOT16}, and included it as the module of motion estimation for two existing state of the art multiple object trackers, \textit{CenterTrack} \cite{zhou2020tracking} and \textit{Tracktor} \cite{bergmann2019tracking}. 
We conduct experiments on the effectiveness of the trajectory estimation comparing both trackers performance with and without trajectory estimation. We use Kalman filter as a trajectory estimator baseline, and compare it to TrajE.
We show that trajectory estimation is useful and, by using TrajE, both tracker performances are improved over two datasets, the MOT challenge (pedestrian) dataset \cite{MOT16}, and the UA-DETRAC (cars in traffic) dataset \cite{CVIU_UA-DETRAC}.

The contributions of this paper are (i) we show empirically that trajectory estimation can be a key factor for a better tracking performance, (ii) we build TrajE, a lightweight model for trajectory estimation based on mixture density networks that can be used as a generic motion model for many trackers, (iii) we add it to two state of the art multiple object trackers, boosting their performance by a considerable margin, and evaluate them against two different datasets on multiple object tracking, setting new state of the art results for CenterTrack + TrajE in both MOTChallenge and UA-DETRAC datasets.

Throughout this paper, we will use the following definitions: \textit{Object ID}, as the identifier assigned to a given object; \textit{Track}, as the list of positions assigned to an Object ID until the current time instant; \textit{Motion}, as the displacement of a given object between two time instants; and \textit{Trajectory}, as a concrete instantiation of the past, current, and future positions that an object can take.

\section{Related Work}
\label{sec:sota}

\textbf{Multiple object tracking}. This task consists on estimating the tracks of multiple objects across a video sequence. Advances in object detection  \cite{redmon2017yolo9000,ren2015faster,liu2016ssd,duan2019centernet} have allowed multiple object trackers to rely on frame-by-frame detections. In consequence, most current algorithms follow the tracking-by-detection paradigm, addressing the tracking problem in two steps: (i) object detection and (ii) association of detections through time to form trajectories.

In tracking, we can differentiate between \textit{offline} \cite{wen2019learning,maksai2017non,schulter2017deep,chen2019aggregate,ma2018customized,tang2017multiple,ma2018trajectory,henschel2017improvements,sheng2018heterogeneous,hornakova2020lifted,braso2020learning,zhang2020long,peng2020tpm} and \textit{online} \cite{bochinski2017high,bergmann2019tracking,chu2019famnet,zhu2018online,sun2019deep,chen2018real,milan2017online,bewley2016simple,wojke2017simple,yang2012multi,chu2019online,chen2017online,xu2019spatial,keuper2018motion,sadeghian2017tracking,feichtenhofer2017detect,zhou2020tracking} methods. Offline methods try to globally optimize the tracks tacking into account a whole video sequence (or a part of it), whereas online methods base their decision on previous observations and current data, and do not take into account future information. In spite of tackling the multiple object tracking problem from different angles, both paradigms have some common challenges to solve. 

To improve object models, \cite{sheng2018heterogeneous,keuper2018motion} combine high-level detections and low-level features (in both cases superpixels) to track the objects, \cite{henschel2017improvements} uses head and body detections for tracking pedestrians, \cite{sun2019deep,chen2017online,ma2018customized} use CNN features on different network depths, while \cite{chen2019aggregate} aggregates multiple features based on temporal attentions, \cite{ijcai2020-74} builds a graph similarity model among objects, and \cite{chu2019famnet,sun2019deep} build deep affinity networks to model the object at different time instants.

To refine tracks \cite{maksai2017non} uses behavioral patterns to achieve global consistency, \cite{wen2019learning} exploits different degrees of dependencies among tracks, \cite{tang2017multiple,zhu2018online,chen2018real,peng2020tpm} focus on filtering noisy detection candidates, \cite{ma2018trajectory} cleaves and reconnects tracks with a siamese recurrent network, \cite{zhang2020long} uses an iterative clustering method to generate tracklets, \cite{hornakova2020lifted,braso2020learning} find a global solution by optimizing a graph, \cite{milan2017online,sadeghian2017tracking} use recurrent neural networks for data association, \cite{chu2019online} encodes awareness within and between target models, and \cite{xu2019spatial} uses relation networks in spatio-temporal domain to combine object cues.

To simplify the tracking pipeline, \cite{bewley2016simple,wojke2017simple} associate detections in $t-1$ to $t$ by means of a Kalman filter and the Hungarian algorithm, \cite{bergmann2019tracking,zhou2020tracking} connect consecutive  detections by regression, \cite{bochinski2017high} directly assumes good detections in all frames and uses the IoU (Intersection over Union) metric to associate the tracks, while \cite{feichtenhofer2017detect} computes the correlation between features from a deep network at multiple depths.

In this work, we propose to use the trajectory information, and present TrajE, a trajectory estimation model that can be added to those aforementioned trackers that use motion information to boost their tracking performance. In addition, the estimated trajectory during an occlusion can be included to the tracks once they are recovered. In this case our method can be considered \textit{offline}, despite no extra computations are done over previous frames.

\noindent \textbf{Trajectory estimation}. This task focuses on modeling and reasoning on the future behavior of agents. 
\cite{rudenko2019human} classifies the different human trajectory prediction methods, based on the model: (i) Physics-based methods, where motion is predicted by dynamics equations based on physical models; (ii) Pattern-based methods, where the dynamics are learned from data; and (iii) Planning-based methods, where there is a reasoning on the agent actions. These models can use (or not) available contextual cues, such as (i) Target agent cues, which are available target agent information; (ii) Dynamic environmental cues, where the target agent is aware of other agents; and (iii) Static environmental cues, where the target agent is aware of the environment information (e.g. static obstacles, such as trees or buildings).


Lately, the trend in the field is to use pattern-based methods. They follow the \textit{Sense - Learn - Predict} paradigm, and learn motion behaviors by fitting different function approximators (i.e. neural networks, hidden Markov models, or Gaussian processes) to data.

To model temporal dependencies, \cite{leal2014learning,pellegrini2010improving,alahi2016social,gupta2018social,leal2011everybody,alahi2017learning,xue2018ss,makansi2019overcoming} combine social information to predict trajectories, \cite{nikhil2018convolutional} uses convolutional neural networks with overlapping windows, \cite{yamaguchi2011you,chen2018real_reinforcement,ren2018collaborative} consider a pedestrian
as an agent that makes decisions to predict the next position, and \cite{ivanovic2019trajectron,salzmann2020trajectron++} define a spatiotemporal graph to predict multiple futures. 

Specifically, \cite{chen2018real_reinforcement,ren2018collaborative} use reinforcement learning to predict the objects position in the next time step, \cite{pellegrini2010improving,leal2014learning,ivanovic2019trajectron,salzmann2020trajectron++} use graphical models to model pedestrian interactions, \cite{makansi2019overcoming} proposes a two-step architecture for overcoming mixture density networks limitations, and \cite{gupta2018social} a GAN for trajectory prediction, using an LSTM to model each trajectory and shares the information as in \cite{alahi2016social}, taking into account obstacles in the scene and other objects in the modelled trajectories.

Similar to TrajE, \cite{alahi2016social,gupta2018social,makansi2019overcoming,ivanovic2019trajectron,salzmann2020trajectron++}, use mixture density networks to predict trajectories. Regarding contextual cues, all these methods are aware of the dynamic obstacles in the form of other agents, and some account for static obstacles in the scene \cite{alahi2016social,gupta2018social}, are aware of the map geometry and topology \cite{makansi2019overcoming} (considering roads and sidewalks in the scene), or are able to use many diverse contextual information \cite{salzmann2020trajectron++}, given its modular nature. All these works use complex models, combining the different agent cues in a social-aware layer in \cite{alahi2016social}, using generative-adversarial networks in \cite{gupta2018social}, introducing the whole image as input to have contextual information in \cite{makansi2019overcoming}, or using variational autoencoders with recurrency in a graph formulation in \cite{ivanovic2019trajectron,salzmann2020trajectron++}.

In contrast, we define TrajE as a much lighter model: it consists on a single layer recurrent neural network, it is unaware of the other agents (dynamic cues) or environmental elements (static cues) in the scene, and the inputs of the network are directly the offsets between the centroids of the object previous positions, i.e. without any other visual or previously ordered information.


To our knowledge, we are the first to bring both the concepts of trajectory estimation using mixture density networks and the posterior trajectory sampling based on beam search to the multiple object tracking framework. 






\section{Trajectory estimation}
\label{sec:trajectory_estimation}

Our model estimates the objects trajectories using a recurrent mixture density network model. Figure \ref{fig:fig_intro} depicts the concept of the trajectory estimator (\textit{TrajE}). Following a set of detections from $t-\tau$ to $t$, our model estimates the parameters of a distribution, in this case a bivariate Gaussian distribution, that models the evolution of the trajectory for the next time step $t+1$. From such a distribution, we sample a set of possible states corresponding to a set of hypotheses for the object position in the next time step.


\subsection{Mixture Density Networks (MDNs)}
\label{subsec:mdn}
Minimization of error functions such as sum-of-squares or cross-entropy leads a neural network model to approximate the conditional averages of the target data, conditioned on the input vector. For problems involving the prediction of continuous variables, the conditional averages provide a very limited information on the underlying structure of data. 

For this reason, predicting the outputs corresponding to input vectors from the conditional probability distribution of the target data might lead to better results, as in handwriting generation \cite{graves2013generating}. To estimate such a distribution we  use Mixture Density Networks (MDNs), proposed in \cite{bishop1994mixture}. These networks combine a conventional neural network with a mixture density model (e.g. a mixture of Gaussians), so that the neural network estimates the distribution parameters. 

Each input vector $x_{t}$ is a real-valued pair $(x_{1,t}, x_{2,t})$, in our case the centroid of a bounding box. The network outputs, $\hat{y}_{t}$, model the different parameters of a distribution $Pr(x_{t+1}|y_{t})$. Some of these outputs model the mixture weights $\pi^{k}_{t}$ while the others estimate the parameters of each mixture component. As we use a mixture of Gaussians, the remaining outputs estimate the mean $\mu_{t}^{k}$, variance $\sigma_{t}^{k}$, and correlation coefficient $\rho_{t}^{k}$ of the $M$ Gaussians that generate the probability distribution. Note that the mean and standard deviation are two dimensional vectors, whereas the weight component and correlation coefficient are scalar.


\begin{equation}
    x_{t} \in \mathbb{R} \times \mathbb{R} 
    \label{eq:obj_pos}
\end{equation}
\begin{equation}
    y_{t} = \{\pi^{k}_{t}, \mu^{k}_{t}, \sigma^{k}_{t}, \rho^{k}_{t}\}_{k=1}^{M}
    \label{eq:mixture_components}
\end{equation}

It is important to state that the mixture components must satisfy several constraints in order to correctly form a valid probability density distribution. To achieve this, several operations are made to the direct outputs of the network $\hat{y}_{t} = \{\hat \pi^{k}_{t}, \hat \mu^{k}_{t}, \hat \sigma^{k}_{t}, \hat \rho^{k}_{t}\}_{k=1}^{M}$. First, weights $\pi^{k}_{t}$ must satisfy 

\begin{equation}
    \sum_{k=1}^{M} \pi^{k}_{t} = 1
    \label{eq:mixture_weights}
\end{equation}

\noindent which can be achieved using the softmax normalization:

\vspace{-1mm}

\begin{equation}
    \pi^{k}_{t} = \frac{\exp(\hat{\pi}^{k}_{t})}{\sum_{k'=1}^{M} \exp(\hat{\pi}^{k'}_{t})}
    \label{eq:softmax}
\end{equation}

For the other components, to keep their values within a meaningful range, $\mu^{k}_{t}$ remains the same, the elements of $\sigma^{k}_{t}$ must be positive, and $\rho^{k}_{t}$ must be a value between $(-1,1)$. Therefore we apply:

\vspace{-1mm}

\begin{equation}
    \mu^{k}_{t} = \hat{\mu}^{k}_{t}; \quad \sigma^{k}_{t} = \exp(\hat{\sigma}^{k}_{t}); \quad \rho^{k}_{t} = \tanh(\hat{\rho}^{k}_{t})
    \label{eq:sigma_exp}
\end{equation}

\vspace{-1mm}

Formally, the probability density $Pr(x_{t+1}|y_{t})$ of the next object position $x_{t+1}$ given the mixture components $y_{t}$ is defined as follows: 

\begin{equation}
    Pr(x_{t+1}|y_{t}) = \sum_{k=1}^{M} \pi_{t}^{k} \mathcal{N}(x_{t+1} | \mu_{t}^{k},\sigma_{t}^{k},\rho_{t}^{k})
    \label{eq:mixture}
\end{equation}

\noindent where

\vspace{-7mm}

\begin{equation}
    \mathcal{N}(x | \mu,\sigma,\rho) = \frac{1}{2 \pi \sigma_{1} \sigma_{2} \sqrt{1-\rho^{2}}} \exp\left [ \frac{-Z}{2(1-\rho^{2})} \right ]
    \label{eq:mixture_gauss}
\end{equation}

\noindent with

\vspace{-6mm}

\begin{equation}
    Z = \frac{(x_{1} - \mu_{1})^{2}}{\sigma_{1}^{2}} + \frac{(x_{2} - \mu_{2})^{2}}{\sigma_{2}^{2}} - \frac{2 \rho (x_{1} - \mu_{1})(x_{2} - \mu_{2})}{\sigma_{1} \sigma_{2}} 
    \label{eq:mixture_z}
\end{equation}

Mixture density networks are trained by maximising the log probability density of the targets under the induced distributions. Equivalently, it is common to minimize the negative log likelihood as loss function.  

\vspace{-5mm}

\begin{equation}
    \mathcal{L}(\mathbf{x}) = \sum_{t=1}^{T} -\log \left (\sum_{k} \pi_{t}^{k} \mathcal{N}(x_{t+1} | \mu_{t}^{k}, \sigma_{t}^{k},\rho_{t}^{k}) \right )
    \label{eq:mixture_loss}
\end{equation}

\noindent where $T$ corresponds to the length of a sequence (in this case an object trajectory).

\vspace{1mm}
\textbf{Biased sampling}. Directly sampling from the generated distribution leads to a huge space of possible trajectories per object. As the process is iterative (an output probability depends on all previous outputs) and the number of beams limited, the generated trajectories may not lead to an optimal solution, as seen in the experiments (Section \ref{subsec:experiments}: Figure \ref{plot:centertrack_var} and Figure \ref{plot:tracktor_var}, for bias $b=0$). A way to palliate this is biasing the sampler towards more likely predictions at each time step. To do so, we modify the distribution as in \cite{graves2013generating}:

\begin{equation}
    \pi^{k}_{t} = \frac{\exp(\hat{\pi}^{k}_{t}(1+b))}{\sum_{k'=1}^{M} \exp(\hat{\pi}^{k'}_{t}(1+b))}
    \label{eq:bias_pi}
\end{equation}

\begin{equation}
    \sigma^{k}_{t} = \exp\left ( \hat{\sigma}^{k}_{t} - b \right )
    \label{eq:bias_sigma}
\end{equation}

\noindent where $b \ge 0$, being Equation \ref{eq:softmax} and Equation \ref{eq:sigma_exp} a specific case for $b=0$. 

\subsection{Trajectory Estimation with MDNs}
\label{subsec:trajectory_estimation}
In this work we estimate an object trajectory for the purpose of multiple object tracking. 
We combine a recurrent mixture density network, as in speech or handwriting synthesis \cite{schuster2000better,graves2013generating}, that aims to model the distribution of the trajectory in time $t+1$, with a beam search technique to explore several track hypotheses. For this problem, using recurrent neural networks is suitable, as the output distribution is conditioned not only on the current input, but on the history of previous inputs.  


To estimate the trajectory, we use the centroid of the detected objects bounding boxes. For each centroid $c^{j}_{t} = (c^{j}_{x,t}, c^{j}_{y,t})$, we use an instance of the trajectory estimator to
estimate a mixture of Gaussians, modeling the distribution from where the centroid $c^{j}_{t+1}$ is going to be sampled. 

Concretely, we feed the difference between the centroids $c^{j}_{t-1}$ and $c^{j}_{t}$, $\Delta(c^{j}_{t-1}, c^{j}_{t})$, to the network and sample the offset in the next time step $\Delta(c^{j}_{t}, c^{j}_{t+1})$ from the distribution. Predicting the trajectory distribution using the offset instead of the position values can be seen as predicting the motion distribution itself. Also, as in \cite{graves2013generating}, using offsets was essential to train the network. For simplicity, we refer to centroids instead of centroids offset for the rest of the paper.

\textbf{Model architecture}. As the number of objects in a scene can grow very large, we use a lightweight architecture that will be instanced per object to track. This architecture is a single layer Gated Recurrent Unit (GRU) \cite{cho2014learning} with a hidden state of size 64 connected with 4 heads of fully connected layers. The two heads that predict the mixture weights $\pi$ and correlation coefficient $\rho$ have dimension $M$, while the two heads that predict the mean $\mu$ and variance $\sigma$ of the coordinates have dimension $2M$. In this work, we set the number of Gaussians in the mixture model to $M=5$. 


\textbf{Beam search}. Beam search is a well known technique in the Natural Language Processing (NLP) community, widely used in tasks like machine translation. When translating a sentence to another language, there are many possible combinations of words that could be outputted, but some solutions are better than others. These solutions may come not in the first word, but after a few generated words. Ideally, keeping all the possible combinations would be optimal in terms of translation, but unpractical in terms of computation. To deal with it, a subset of options is used and explored in order to find the best translation to the given text.

We use beam search to take into account multiple trajectory hypotheses, as sampling a single point (even the most likely one) is prone to drifting. In beam search, $B$ stands for beam width. In machine translation the top-$B$ predicted words are considered at every estimation step. 


In our case, $B$ encodes both the beam width and the amount of centroid positions we are going to sample from the trajectory distribution. We start with $B$ hypotheses for the object location in the next time step $t+1$. These hypotheses are propagated and, if there is no detection, again $B$ hypotheses are generated, ending up with $B^{2}$ hypotheses. To avoid an exponential growth, we prune to keep $B$ trajectory hypotheses, as in machine translation algorithms. Note that, for $B=1$, beam search becomes the greedy search algorithm. In TrajE, this cycle is over when a) the trajectory ID is assigned to a detection, or b) the track patience counter is over, and the track is terminated.  

\section{Multiple Object Tracking using MDNs}
\label{sec:mot}

\begin{figure*}[t]
    \centering
    \includegraphics[width=\textwidth]{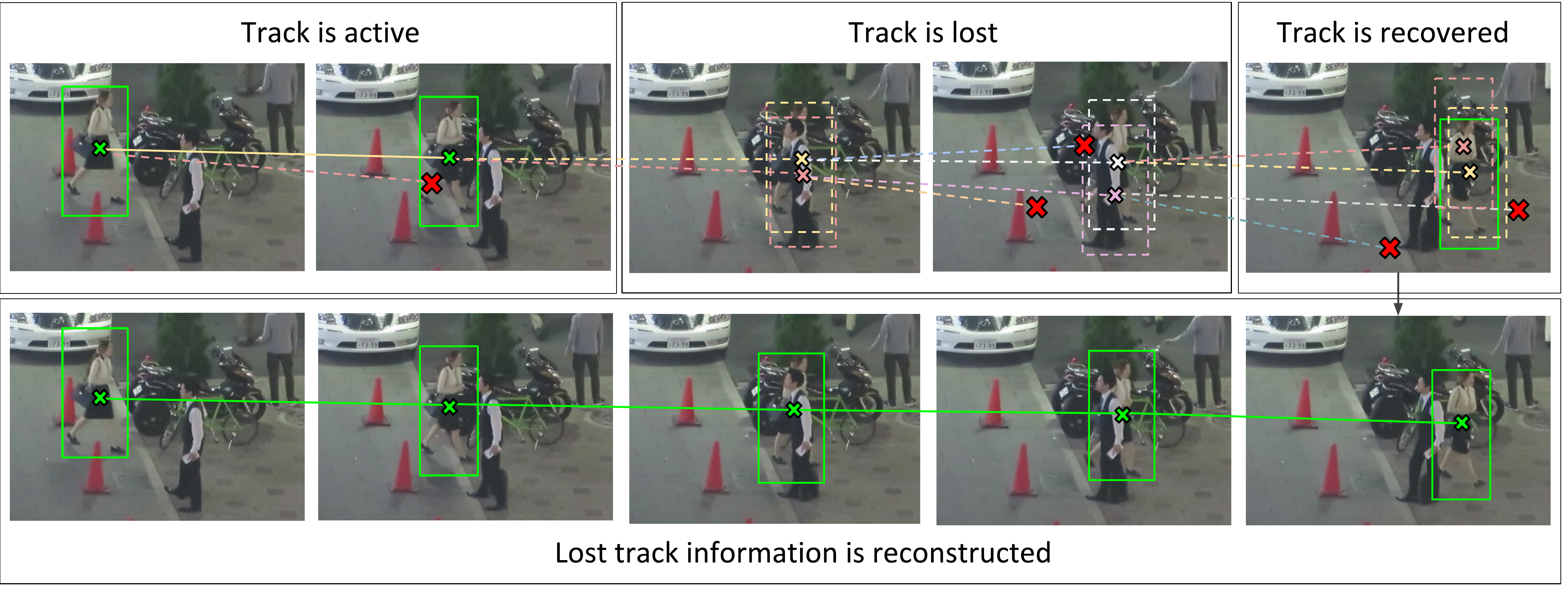}
    \caption{Visualization of the beam search and occlusion handling. First, $B$ (for illustration purposes in this case $B=2$) hypotheses of where the object could be in the next time step are sampled from the generated trajectory distribution. If a detection is associated to a track, the beam search corresponding to that track is reset. If no detection is associated to an active track, several hypotheses are sampled from the trajectory distribution and pruned (red crosses) in the next time step to keep $B$ hypotheses. If the track is recovered (a detection is assigned to the track again), and the estimated trajectory is coherent with it, the computed trajectory during the lost state of the track is added to the object track. The occlusion-filling bounding boxes are generated using the centroids of the estimated trajectory, and the width and height of the new detection.}
    \label{fig:occ_handling}
\end{figure*}

We integrated our trajectory estimation model, \textit{TrajE}, into two existing multiple object trackers, \textit{CenterTrack} \cite{zhou2020tracking} and \textit{Tracktor} \cite{bergmann2019tracking}, replacing their original motion estimation modules. In this section we cover the integration of TrajE into both trackers, track handle policies, and experiments.



\textbf{Baselines.} CenterTrack and Tracktor are defined as trackers-by-regression, meaning that, by "regressing" the detections in $t-1$ to detections in $t$, they are able to assign the newly detected object in $t$ to the same track where the original object in $t-1$ belonged.

The main idea in CenterTrack is to regress consecutive frame centroids to associate the tracked objects from $t-1$ to the detections in $t$. They extend the CenterNet \cite{duan2019centernet} object detector to a network able to associate detections over time by means of feeding the model two consecutive images ($I_{t-1}$, $I_{t}$) and the heatmap of the detections in the previous frame ($\mathcal{H}_{t-1}$), biasing the detector towards previous detections. To further boost their performance, the model computes, as the motion model, the offset between detections in two consecutive time steps, as some sort of sparse optical flow between object centroids. By modeling the objects as points and estimating their motion as offsets, the tracks and detections are associated by means of L1 distance.

Tracktor also forms tracks by propagating tracked objects from $t-1$ to $t$, regressing them using the ROI-pooling layer present in Faster-RCNN \cite{ren2015faster}, as if they were object proposals computed by the network. To improve their results, they make use of re-identification using siamese networks and a motion model based on the Constant Velocity (CV) assumption or Camera Motion Compensation (CMC), depending on whether there is large camera movement in the video sequence or not.

Both methods heavily rely on the positioning of the objects to associate them with the existing tracks projected from $t-1$, leading to the intuition that a better object projection between time instants should directly boost the trackers performance. Also, with a reliable estimation of the trajectory, the re-identification and occlusion reconstruction problems would be reduced, as both trackers association mechanism is highly biased towards the detections on the previous frame, making it difficult to associate lost objects further in time (from $t-\tau$) to newly detected objects in $t$.


\subsection{Trajectory estimation applied to tracking}


Each object being tracked has an \textit{active} or \textit{lost} state, depending on whether it is detected and associated to an existing track, or lost due to occlusions, false negatives in the detection phase, or out of the scene. 

When a track is initialized, a patience timer is assigned to it. For every frame the track is not associated to any detection (lost state), the patience value of the track is decreased by $1$, until it becomes $0$, when that object track will be considered as terminated. If the object is associated to a track (active state) the patience for that specific track is reset. As TrajE helps the tracker handle occlusions, we set this patience value to $100$, allowing objects to reappear in the short-mid term. To add TrajE to a tracker, we define the following common track handle policies:  


\noindent\textbf{Active tracks.} An active track in $t-1$, with its $B$ position hypotheses in $t$, is associated with a detection in $t$. The beam corresponding to the trajectory that best fits the detection is chosen, its hidden state copied to the other $B-1$ instances of the trajectory estimator, and its patience reset. The input to the trajectory estimator will be the offset between the new detection and the previous one, and there will be again $B$ possible object mappings in $t+1$.


\noindent\textbf{Lost tracks.} If a track in $t$ is not associated to any detection, the track will be considered as lost. From this moment until a detection is associated with this track (using re-identification) in $t + \tau$, the track will be in the lost state. The $B$ sampled hypotheses in $t-1$ are kept, and fed to $B$ new instances of the trajectory estimation network in $t$. The resulting $B^{2}$ hypotheses are pruned to $B$ to avoid exponential growth.
Forwarding the information of lost tracks is important to re-identify the track and recover from occlusions. Also, the object track after an occlusion can be reconstructed if the new detection is coherent with the estimated trajectory.




\noindent\textbf{Recovered tracks.} If a lost track since $t-\tau$ is associated to a detection in $t$, the best beam given the likelihood of the estimated trajectory is associated to that track, the beam search exploration is reset, and a decision is taken whether the trajectory estimated during the lost state (e.g. due to an occlusion) is added to the track or not.

\noindent This decision is made regarding the spatio-temporal coherence of the new detection and the estimated trajectory. To compare both, we consider a bounding box with the centroid of the estimated trajectory and the width and height of the last detection in $t - \tau$. If the detection associated to the track in $t$ and the estimated trajectory of the lost object in $t - \tau$ have an Intersection over Union (IoU) above a certain threshold (we use $0.5$), we consider that TrajE generated a good trajectory estimation, and keep the trajectory as part of the path that the track has followed. Otherwise, we discard the estimated trajectory during the occlusion, and associate only the new detection to the track. 

\noindent\textbf{Terminated tracks.} If the track reaches the maximum patience in the lost state, it will be terminated and removed from the possible re-identification with new detections.


\subsection{Exploration strategies}
\label{subsec:MOT_trackers}
Once the trajectory distribution is computed, we propose three approaches to integrate TrajE to an object tracker based on the exploration strategy. The first one is the Best Mean (BM), which takes the most likely mean of one of the Gaussians forming the probability distribution of the trajectory.
The second one is a Greedy Beam Search (GBS). It takes the local best sample (using maximum likelihood) at every time step given the estimated distribution as the motion of the tracked object. Note that the difference between GBS and BM strategies is that GBS samples from the distribution, while BM assumes that the best possible position in $t+1$ is the mean of the Gaussian with highest likelihood of the mixture of Gaussians.
The last strategy is a Pure Beam Search (PBS) that uses all $B$ hypotheses to forward the track in $B$ possible ways. If a detection is associated to the track, the best beam given the historic is chosen in order to keep a single detection per track at every time step. Note that, for $B=1$, both GBS and PBS strategies become the same.

For CenterTrack, we swapped its original offset estimation directly with BM, GBS, or PBS. Note that CenterTrack's offset estimation matches detections by estimating the objects offset from $t$ to $t-1$, while TrajE estimates the trajectory of the objects from $t-1$ to $t$.
For BM, we swapped the motion module output (offset estimation in this case) with the mean of the Gaussian with highest likelihood in the distribution generated by TrajE. In a similar way, for the GBS strategy we take the the best sample given by TrajE at every time step. 
For the PBS, we take the $B$ position hypotheses per object in $t$, compare them to the detections of the CenterTrack in $t$, and solve an assignation problem to end up with their closest detection (if any). For all the different exploration strategies, the assignment of IDs to the detections follows the same strategy as in CenterTrack, which links the projected object from $t-1$ to $t$ to its closest detection in $t$, in terms of L1 distance.

Tracktor relies on the regression from the Faster-RCNN's \textit{ROI pooling} layer to map the objects from $t-1$ to $t$. 
For the BM and GBS strategies we follow the same strategy as in CenterTrack's case, i.e. swapping the motion estimation by the next most likely mean or sample (for BM and GBS respectively), projecting the object to its next probable position in $t$.
For the PBS, $B$ hypotheses are projected to the next time step (using the width and height of the projected bounding box), and $B$ regressions per track are made. If several detections are associated to the track, the most probable trajectory associated to a detection (given maximum likelihood) is chosen.

\subsection{Experiments}
\label{subsec:experiments}

Experiments and incremental studies are performed in the MOTChallenge dataset \cite{MOT16} (pedestrians). To see whether adding TrajE to a tracker generalizes, we also evaluate it against the UA-DETRAC dataset \cite{CVIU_UA-DETRAC} (cars), using the same trajectory estimation model trained on pedestrians from the MOTChallenge training set. 

The MOTChallenge dataset contains 14 challenging video sequences recording pedestrians (7 for training, 7 for testing) with both static and moving camera, recorded in the wild (unconstrained environments), and different locations. The UA-DETRAC dataset consists on 100 videos (60 for training, 40 for testing) with static camera (stable or unstable), recording the road traffic. 

For the MOTChallenge, we use the object detection models provided by the two trackers (and re-identification in case of Tracktor), for the UA-DETRAC dataset we re-trained those models for car detection and re-identification.

For a fair comparison, we use both trackers in their public detection mode, as MOTChallenge requires the provided (public) detections to be used. Both trackers extend their tracking algorithm with their own detector to the public detection setting by initializing the tracks only when a public detection is present.

\noindent\textbf{Training.} To train TrajE we used the MOT17 \cite{MOT16} challenge data. To generate the training data, we sampled random trajectories from objects in different sequences. These trajectories are split in batches of $100$ points in order to have different beginnings and ends of a trajectory. Also, we apply noise to the input sequences to make our model more robust to noisy detections. We end up with $20000$ sequences for training and $2000$ as validation set. We trained the model for $100$ epochs with a learning rate of $0.001$, multiplying it by a learning decay of $0.1$ at iterations $15$, $40$, $80$. 

\noindent \textbf{Incremental study.}  Table \ref{tab:ablation_centertrack} and Table \ref{tab:ablation_tracktor}  present a study on whether  CenterTrack and Tracktor benefit from estimating the object trajectories. In this study we include popular metrics in MOT challenges: MOTA, IDF1, the recently introduced HOTA metric \cite{luiten2020hota}, which balances the tracking score, both in detection (DetA) and association (AssA) terms, and the identity switches (IDSW) between tracks.

\begin{table}
\center
\tabcolsep=0.11cm
    \resizebox{\columnwidth}{!}{
    \begin{tabular}{l c c c | c c | c}\\
     \toprule
      & MOTA $\uparrow$ & IDF1 $\uparrow$ & HOTA $\uparrow$ & DetA $\uparrow$ & AssA $\uparrow$ & IDSW $\downarrow$ \\ [0.5ex] 
     
     \midrule
     
     CenterTrack & 67.4 & 62.7 & 63.0 & 69.3 & 57.2 & 1356\\
     
     \midrule
     +OFF & 67.7 &  63.8 & 63.8 & 69.2 & 58.8 & 1077\\
     
     \midrule
     +Kalman & 68.7 &  64.4 & 64.0 & 70.9 & 57.7 & 1642\\
     +BM & 68.4 &  65.6 & 64.9 & 69.9 & 60.2  & 833\\
     +GBS & 68.9 $\pm$ 0.3 &  65.6 $\pm$ 0.5 & 65.1 $\pm$ 0.5 & 70.3 $\pm$ 0.3 & 60.2 $\pm$ 1.0 & 857\\
     +PBS & 69.0 $\pm$ 0.2 &  66.0 $\pm$ 0.1 & 65.3 $\pm$ 0.2 & 70.4 $\pm$ 0.1 & 60.5 $\pm$ 0.4 & 789\\
     
     \midrule     
     +Kalman+occ & 68.8 &  64.6 & 64.1 &  \textbf{71.2} &  57.8 & 1382\\
     +BM+occ & 69.1 &  65.9 & 65.4 & 70.6 &  60.5  & 732\\
     +GBS+occ & 69.5 $\pm$ 0.2 & 66.0 $\pm$ 0.5 & 65.5 $\pm$ 0.5 &  70.9 $\pm$ 0.2 & 60.6 $\pm$ 1.1 & 751\\
     +PBS+occ & \textbf{69.6} $\pm$ 0.1  &  \textbf{66.3} $\pm$ 0.1 & \textbf{65.7} $\pm$ 0.2 & 71.0 $\pm$ 0.1 & \textbf{60.9} $\pm$ 0.4 & \textbf{706}\\
     
     \midrule
     
    \end{tabular}}
\vspace{0.2mm}
\caption{Incremental study on CenterTrack+TrajE with the exploration strategies, GBS, PBS, and BM, and CenterTrack+Kalman filter for trajectory estimation. For PBS and GBS we depict the mean and variance over 5 runs. OFF stands for offset prediction present in CenterTrack, and occ for occlusion reconstruction from trajectories. We compare the performance on the full MOT17 training set using the Faster R-CNN public detections, with $bias=1$, $B=5$.}

\label{tab:ablation_centertrack}
\vspace{-3mm}
\end{table}

\begin{table}
\center
\tabcolsep=0.11cm
    \resizebox{\columnwidth}{!}{
    \begin{tabular}{l c c c | c c | c}\\
     \toprule
     & MOTA $\uparrow$ & IDF1 $\uparrow$ & HOTA $\uparrow$ & DetA $\uparrow$ & AssA $\uparrow$ & IDSW $\downarrow$ \\ [0.5ex] 
      
     \midrule
     Tracktor v2 & 60.6 & 62.0 & 60.7 & 61.6 & 59.7 & 913 \\
     \midrule
     +CV & 61.2 &  63.9 & 62.0 & 61.9 &  62.1 & 557\\
     +CV+CMC & 61.7 &  64.9 & 62.8 &  62.1 &  63.5 & \textbf{269}\\
     
     \midrule
     +Kalman & 61.4 &  65.2 & 63.1 &  62.0 & 64.2 & 462\\
     +BM & 61.6 & 66.4 & 63.7 & 62.1 &  65.3 & 399 \\
     +GBS & 61.4 $\pm$ 0.1 & 66.6 $\pm$ 0.5 & 63.6 $\pm$ 0.3 & 62.0 $\pm$ 0.1 & 65.4 $\pm$ 0.6 & 407\\
     +PBS & 61.5 $\pm$ 0.1 & 66.7 $\pm$ 0.5 & 63.7 $\pm$ 0.3 & 62.1 $\pm$ 0.1 & 65.5 $\pm$ 0.6 & 369\\

     \midrule 
     +Kalman+occ & 61.2 &  66.9 & \textbf{64.4} &  \textbf{62.8} &  \textbf{66.2}  & 506\\
     +BM+occ & \textbf{62.2} & 66.8 & 64.2 & \textbf{62.8} & 65.7  & 399\\
     +GBS+occ & 61.9 $\pm$ 0.1 & 66.9 $\pm$ 0.5 & 64.1 $\pm$ 0.3 & 62.5 $\pm$ 0.1 & 65.7 $\pm$ 0.6 & 409\\
     +PBS+occ & 62.0  $\pm$ 0.1 & \textbf{67.0}  $\pm$ 0.4 & 64.2 $\pm$ 0.3 & 62.6 $\pm$ 0.1 & 65.8 $\pm$ 0.6 & 370\\

     \midrule
    \end{tabular}}
\vspace{0.05mm}
\caption{Incremental study on Tracktor+TrajE with the exploration strategies, GBS, PBS, and BM, and Tracktor+Kalman filter for trajectory estimation. For PBS and GBS we depict the mean and variance over 5 runs. CV stands for Constant Velocity assumption, CMC for Camera Motion Compensation, and occ for occlusion reconstruction from trajectories. We compare the performance on the full MOT17 training set using the Faster R-CNN public detections, with $bias=1$, $B=5$.}

\label{tab:ablation_tracktor}
\vspace{-5mm}
\end{table}

The starting point are two baselines: the trackers without and with their own defined motion estimation (see the baselines in the beginning of Section \ref{sec:mot} for details). We then swap this motion estimation for TrajE, and study the impact of the three exploration strategies, Best Mean (BM), Greedy Beam Search (GBS), and Pure Beam Search (PBS) with and without occlusion reconstruction using the estimated trajectories. To see whether the benefits of including trajectory information into object trackers generalize, we adapt the Kalman filter implemented in SORT \cite{bewley2016simple} for trajectory estimation, by generating trajectories for a track whose state is set to lost, while including the new detections assigned to the track whenever its state is set to active.

\noindent \textbf{Motion estimation.} The first observation is that any provided motion information is helpful to the trackers. In Table \ref{tab:ablation_centertrack}, by computing the motion of the object with an offset head for CenterTrack, and Table  \ref{tab:ablation_tracktor}, by assuming a constant velocity (CV) of an object for Tracktor, both trackers improve their performance considerably. 

\noindent \textbf{Trajectory estimation.} Following the incremental study, we observe that, by adding TrajE, both trackers consistently improve in all metrics. 
As TrajE has a sampling step, the mean and variance for five runs is depicted. Comparing both TrajE and the Kalman filter extended to trajectories, TrajE provides much more gain to CenterTrack in Table \ref{tab:ablation_centertrack}, and a similar gain to Tracktor in Table \ref{tab:ablation_tracktor}. This difference in the increment is due to the way both trackers treat the regression step. While CenterTrack computes directly the distance between the projected detection in $t$ with the actual detection in $t$, Tracktor regresses the projected detection in $t$ with the ROI-pooling layer present in the Faster-RCNN, conditioning the trajectory estimation to the object detector, leading to similar solutions if the projected detections of TrajE and the Kalman filter are close to each other. Also, by reconstructing the occlusions in the tracks by using the estimated trajectories, there is an incremental gain for all the trajectory estimation strategies.

\noindent \textbf{TrajE parameters.}  Figure \ref{plot:centertrack_var} and Figure \ref{plot:tracktor_var} study the impact of the \textit{bias} and beam width (\textit{B}) values for the different trajectory estimation strategies. We show the behavior of the two parameters over the different exploration strategies, BM, GBS, and PBS, and compare them with the Kalman filter adapted for trajectory estimation, and the trackers baseline (using their own motion estimators). In the trajectory estimation strategies, the occlusion reconstruction is used. Note that for $B=1$, GBS and PBS strategies are the same.

The results show how, by using beam search, the trackers benefit from the several hypotheses, and how biasing the outputs of the MDN can become crucial in the trajectory estimation. Regarding the exploration strategies, PBS has better overall results than GBS. Interestingly, by using the BM strategy, which is equivalent to use a beam width of $B=1$ with $bias\to\infty$, TrajE also boosts the performance of the trackers by a considerable margin, and can be a good alternative when faster computation is mandatory.

\begin{figure}
    \begin{minipage}[b]{\linewidth}
        \centering
        \pgfplotsset{compat = 1.3}
\pgfplotstableread{plots/bw_1/centertrack_pbs_OVERALL.tex}\varpbsone
\pgfplotstableread{plots/bw_3/centertrack_pbs_OVERALL.tex}\varpbsthree
\pgfplotstableread{plots/bw_5/centertrack_pbs_OVERALL.tex}\varpbsfive
\pgfplotstableread{plots/bw_10/centertrack_pbs_OVERALL.tex}\varpbsten
\pgfplotstableread{plots/bw_1/centertrack_gbs_OVERALL.tex}\vargbsone
\pgfplotstableread{plots/bw_3/centertrack_gbs_OVERALL.tex}\vargbsthree
\pgfplotstableread{plots/bw_5/centertrack_gbs_OVERALL.tex}\vargbsfive
\pgfplotstableread{plots/bw_10/centertrack_gbs_OVERALL.tex}\vargbsten

\begin{center}
\begin{minipage}{\linewidth}
\resizebox{0.78\linewidth}{!}{\begin{tikzpicture}[/pgfplots/width=\linewidth, /pgfplots/height=\linewidth, /pgfplots/legend pos=south east]
\begin{axis}[
    	ymin=66.5, ymax=70.5,
        xmin=0, xmax=5,
        ylabel={\Large MOTA},
        title={\Large $B=1$},
        xlabel=Bias,
		font=\Large,
        grid=both,
		grid style=dotted,
        xmode=linear,
        ytick={67, 68, 69, 70},
		yticklabels={67, 68, 69, 70},
        xtick={0, 1, 2, 3, 4, 5},
	    xticklabels={0, 0.1, 0.5, 1, 5, 10},
        enlarge x limits=0.05,
        legend style={at={(2.6,1.2)}, /tikz/every even column/.append style={column sep=3.5mm}},
        legend columns = -1,
        ]
        \addlegendentry{CenterTrack+OFF}
		\addplot[color=c_centertrack,mark=*, mark size=1, line width=2, dotted]
		coordinates{(0,67.7)(1,67.7)(2,67.7)(3,67.7)(4,67.7)(5,67.7)};
        \addlegendentry{Kalman}
		\addplot[color=c_kalman,mark=*, mark size=1, line width=2, dotted]
		coordinates{(0,68.8)(1,68.8)(2,68.8)(3,68.8)(4,68.8)(5,68.8)};
		\addlegendentry{BM}
		\addplot[color=c_best_mean,mark=*, mark size=1, line width=2, dotted]
		coordinates{(0,69.1)(1,69.1)(2,69.1)(3,69.1)(4,69.1)(5,69.1)};
		
		\addlegendentry{GBS}
		\addplot[color=c_vgbs,mark=*, mark size=1, line width=2] table[x=num,y=MOTA_mean] {\vargbsone};
		\addlegendentry{PBS}
		\addplot[color=c_vpbs,mark=*, mark size=1, line width=2] table[x=num,y=MOTA_mean] {\varpbsone};
		
		\addplot[name path=us_top,color=c_vgbs!70] table[x=num,y=MOTA_max] {\vargbsone};
		\addplot[name path=us_down,color=c_vgbs!70] table[x=num,y=MOTA_min] {\vargbsone};
		\addplot[c_vgbs!25,fill opacity=0.5] fill between[of=us_top and us_down];
		\addplot[name path=us_top,color=c_vpbs!70] table[x=num,y=MOTA_max] {\varpbsone};
		\addplot[name path=us_down,color=c_vpbs!70] table[x=num,y=MOTA_min] {\varpbsone};
		\addplot[c_vpbs!25,fill opacity=0.5] fill between[of=us_top and us_down];

    \end{axis}

\begin{axis}[
    	ymin=66.5, ymax=70.5,
        xmin=0, xmax=5,
        title={\Large $B=5$},
        xlabel=Bias,
		font=\Large,
        grid=both,
		grid style=dotted,
        xmode=linear,
        ytick={67, 68, 69, 70},
		yticklabels={67, 68, 69, 70},
        xtick={0, 1, 2, 3, 4, 5},
	    xticklabels={0, 0.1, 0.5, 1, 5, 10},
        enlarge x limits=0.05,
        legend style={ at={(1,0)}},
        xshift=8cm
        ]

		\addplot[color=c_vgbs,mark=*, mark size=1, line width=2] table[x=num,y=MOTA_mean] {\vargbsfive};
		\addplot[color=c_vpbs,mark=*, mark size=1, line width=2] table[x=num,y=MOTA_mean] {\varpbsfive};

		\addplot[color=c_centertrack,mark=*, mark size=1, line width=2, dotted]
		coordinates{(0,67.7)(1,67.7)(2,67.7)(3,67.7)(4,67.7)(5,67.7)};
		\addplot[color=c_kalman,mark=*, mark size=1, line width=2, dotted]
		coordinates{(0,68.8)(1,68.8)(2,68.8)(3,68.8)(4,68.8)(5,68.8)};
		\addplot[color=c_best_mean,mark=*, mark size=1, line width=2, dotted]
		coordinates{(0,69.1)(1,69.1)(2,69.1)(3,69.1)(4,69.1)(5,69.1)};
		
		\addplot[name path=us_top,color=c_vgbs!70] table[x=num,y=MOTA_max] {\vargbsfive};
		\addplot[name path=us_down,color=c_vgbs!70] table[x=num,y=MOTA_min] {\vargbsfive};
		\addplot[c_vgbs!25,fill opacity=0.5] fill between[of=us_top and us_down];
		\addplot[name path=us_top,color=c_vpbs!70] table[x=num,y=MOTA_max] {\varpbsfive};
		\addplot[name path=us_down,color=c_vpbs!70] table[x=num,y=MOTA_min] {\varpbsfive};
		\addplot[c_vpbs!25,fill opacity=0.5] fill between[of=us_top and us_down];

    \end{axis}

\begin{axis}[
    	ymin=66.5, ymax=70.5,
        xmin=0, xmax=5,
        title={\Large $B=10$},
        xlabel=Bias,
		font=\Large,
        grid=both,
		grid style=dotted,
        xmode=linear,
        ytick={67, 68, 69, 70},
		yticklabels={67, 68, 69, 70},
        xtick={0, 1, 2, 3, 4, 5},
	    xticklabels={0, 0.1, 0.5, 1, 5, 10},
        enlarge x limits=0.05,
        legend style={ at={(1,0)}},
        xshift=16cm
        ]

		\addplot[color=c_vgbs,mark=*, mark size=1, line width=2] table[x=num,y=MOTA_mean] {\vargbsten};
		\addplot[color=c_vpbs,mark=*, mark size=1, line width=2] table[x=num,y=MOTA_mean] {\varpbsten};
        
		\addplot[color=c_centertrack,mark=*, mark size=1, line width=2, dotted]
		coordinates{(0,67.7)(1,67.7)(2,67.7)(3,67.7)(4,67.7)(5,67.7)};
		\addplot[color=c_kalman,mark=*, mark size=1, line width=2, dotted]
		coordinates{(0,68.8)(1,68.8)(2,68.8)(3,68.8)(4,68.8)(5,68.8)};
		\addplot[color=c_best_mean,mark=*, mark size=1, line width=2, dotted]
		coordinates{(0,69.1)(1,69.1)(2,69.1)(3,69.1)(4,69.1)(5,69.1)};
		
		\addplot[name path=us_top,color=c_vgbs!70] table[x=num,y=MOTA_max] {\vargbsten};
		\addplot[name path=us_down,color=c_vgbs!70] table[x=num,y=MOTA_min] {\vargbsten};
		\addplot[c_vgbs!25,fill opacity=0.5] fill between[of=us_top and us_down];
		\addplot[name path=us_top,color=c_vpbs!70] table[x=num,y=MOTA_max] {\varpbsten};
		\addplot[name path=us_down,color=c_vpbs!70] table[x=num,y=MOTA_min] {\varpbsten};
		\addplot[c_vpbs!25,fill opacity=0.5] fill between[of=us_top and us_down];

    \end{axis}

\end{tikzpicture}}
\end{minipage}
\end{center}
    \end{minipage}
    \begin{minipage}[b]{\linewidth}
        \centering
        \pgfplotsset{compat = 1.3}
\pgfplotstableread{plots/bw_1/centertrack_pbs_OVERALL.tex}\varpbsone
\pgfplotstableread{plots/bw_3/centertrack_pbs_OVERALL.tex}\varpbsthree
\pgfplotstableread{plots/bw_5/centertrack_pbs_OVERALL.tex}\varpbsfive
\pgfplotstableread{plots/bw_10/centertrack_pbs_OVERALL.tex}\varpbsten
\pgfplotstableread{plots/bw_1/centertrack_gbs_OVERALL.tex}\vargbsone
\pgfplotstableread{plots/bw_3/centertrack_gbs_OVERALL.tex}\vargbsthree
\pgfplotstableread{plots/bw_5/centertrack_gbs_OVERALL.tex}\vargbsfive
\pgfplotstableread{plots/bw_10/centertrack_gbs_OVERALL.tex}\vargbsten

\begin{center}
\begin{minipage}{\textwidth}
\resizebox{0.336\linewidth}{!}{\begin{tikzpicture}[/pgfplots/width=\linewidth, /pgfplots/height=\linewidth, /pgfplots/legend pos=south east]
\begin{axis}[
    	ymin=60.5, ymax=68.5,
        xmin=0, xmax=5,
        ylabel={\Large IDF1},
        xlabel=Bias,
		font=\Large,
        grid=both,
		grid style=dotted,
        xmode=linear,
        ytick={61, 62, 63, 64, 65, 66, 67, 68},
		yticklabels={61, 62, 63, 64, 65, 66, 67, 68},
        xtick={0, 1, 2, 3, 4, 5},
	    xticklabels={0, 0.1, 0.5, 1, 5, 10},
        enlarge x limits=0.05,
        ]
		\addplot[color=c_vgbs,mark=*, mark size=1, line width=2] table[x=num,y=IDF1_mean] {\vargbsone};
		\addplot[color=c_vpbs,mark=*, mark size=1, line width=2] table[x=num,y=IDF1_mean] {\varpbsone};

		\addplot[color=c_centertrack,mark=*, mark size=1, line width=2, dotted]
		coordinates{(0,63.8)(1,63.8)(2,63.8)(3,63.8)(4,63.8)(5,63.8)};
		\addplot[color=c_kalman,mark=*, mark size=1, line width=2, dotted]
		coordinates{(0,64.6)(1,64.6)(2,64.6)(3,64.6)(4,64.6)(5,64.6)};
		\addplot[color=c_best_mean,mark=*, mark size=1, line width=2, dotted]
		coordinates{(0,65.9)(1,65.9)(2,65.9)(3,65.9)(4,65.9)(5,65.9)};
		
		\addplot[name path=us_top,color=c_vgbs!70] table[x=num,y=IDF1_max] {\vargbsone};
		\addplot[name path=us_down,color=c_vgbs!70] table[x=num,y=IDF1_min] {\vargbsone};
		\addplot[c_vgbs!25,fill opacity=0.5] fill between[of=us_top and us_down];
		\addplot[name path=us_top,color=c_vpbs!70] table[x=num,y=IDF1_max] {\varpbsone};
		\addplot[name path=us_down,color=c_vpbs!70] table[x=num,y=IDF1_min] {\varpbsone};
		\addplot[c_vpbs!25,fill opacity=0.5] fill between[of=us_top and us_down];

    \end{axis}

\begin{axis}[
    	ymin=60.5, ymax=68.5,
        xmin=0, xmax=5,
        xlabel=Bias,
		font=\Large,
        grid=both,
		grid style=dotted,
        xmode=linear,
        ytick={61, 62, 63, 64, 65, 66, 67, 68},
		yticklabels={61, 62, 63, 64, 65, 66, 67, 68},
        xtick={0, 1, 2, 3, 4, 5},
	    xticklabels={0, 0.1, 0.5, 1, 5, 10},
        enlarge x limits=0.05,
        xshift=8cm
        ]

		\addplot[color=c_vgbs,mark=*, mark size=1, line width=2] table[x=num,y=IDF1_mean] {\vargbsfive};
		\addplot[color=c_vpbs,mark=*, mark size=1, line width=2] table[x=num,y=IDF1_mean] {\varpbsfive};

		\addplot[color=c_centertrack,mark=*, mark size=1, line width=2, dotted]
		coordinates{(0,63.8)(1,63.8)(2,63.8)(3,63.8)(4,63.8)(5,63.8)};
		\addplot[color=c_kalman,mark=*, mark size=1, line width=2, dotted]
		coordinates{(0,64.6)(1,64.6)(2,64.6)(3,64.6)(4,64.6)(5,64.6)};
		\addplot[color=c_best_mean,mark=*, mark size=1, line width=2, dotted]
		coordinates{(0,65.9)(1,65.9)(2,65.9)(3,65.9)(4,65.9)(5,65.9)};
		
		\addplot[name path=us_top,color=c_vgbs!70] table[x=num,y=IDF1_max] {\vargbsfive};
		\addplot[name path=us_down,color=c_vgbs!70] table[x=num,y=IDF1_min] {\vargbsfive};
		\addplot[c_vgbs!25,fill opacity=0.5] fill between[of=us_top and us_down];
		\addplot[name path=us_top,color=c_vpbs!70] table[x=num,y=IDF1_max] {\varpbsfive};
		\addplot[name path=us_down,color=c_vpbs!70] table[x=num,y=IDF1_min] {\varpbsfive};
		\addplot[c_vpbs!25,fill opacity=0.5] fill between[of=us_top and us_down];

    \end{axis}

\begin{axis}[
    	ymin=60.5, ymax=68.5,
        xmin=0, xmax=5,
        xlabel=Bias,
		font=\Large,
        grid=both,
		grid style=dotted,
        xmode=linear,
        ytick={61, 62, 63, 64, 65, 66, 67, 68},
		yticklabels={61, 62, 63, 64, 65, 66, 67, 68},
        xtick={0, 1, 2, 3, 4, 5},
	    xticklabels={0, 0.1, 0.5, 1, 5, 10},
        enlarge x limits=0.05,
        xshift=16cm
        ]

		\addplot[color=c_vgbs,mark=*, mark size=1, line width=2] table[x=num,y=IDF1_mean] {\vargbsten};
		\addplot[color=c_vpbs,mark=*, mark size=1, line width=2] table[x=num,y=IDF1_mean] {\varpbsten};

		\addplot[color=c_centertrack,mark=*, mark size=1, line width=2, dotted]
		coordinates{(0,63.8)(1,63.8)(2,63.8)(3,63.8)(4,63.8)(5,63.8)};
		\addplot[color=c_kalman,mark=*, mark size=1, line width=2, dotted]
		coordinates{(0,64.6)(1,64.6)(2,64.6)(3,64.6)(4,64.6)(5,64.6)};
		\addplot[color=c_best_mean,mark=*, mark size=1, line width=2, dotted]
		coordinates{(0,65.9)(1,65.9)(2,65.9)(3,65.9)(4,65.9)(5,65.9)};
		
		\addplot[name path=us_top,color=c_vgbs!70] table[x=num,y=IDF1_max] {\vargbsten};
		\addplot[name path=us_down,color=c_vgbs!70] table[x=num,y=IDF1_min] {\vargbsten};
		\addplot[c_vgbs!25,fill opacity=0.5] fill between[of=us_top and us_down];
		\addplot[name path=us_top,color=c_vpbs!70] table[x=num,y=IDF1_max] {\varpbsten};
		\addplot[name path=us_down,color=c_vpbs!70] table[x=num,y=IDF1_min] {\varpbsten};
		\addplot[c_vpbs!25,fill opacity=0.5] fill between[of=us_top and us_down];

    \end{axis}

\end{tikzpicture}}
\end{minipage}
\end{center}
    \end{minipage}
    \caption{Experiments over TrajE parameters $bias$ and beam width $B$ for CenterTrack using TrajE or Kalman with occlusion reconstruction. The y-axis corresponds to the metric score, the x-axis to the $bias$, and the columns to the different $B$ values. 
    Solid lines correspond to the average value over five runs, and the transparent region limits correspond to the maximum and minimum values.}
    \label{plot:centertrack_var}
\end{figure}
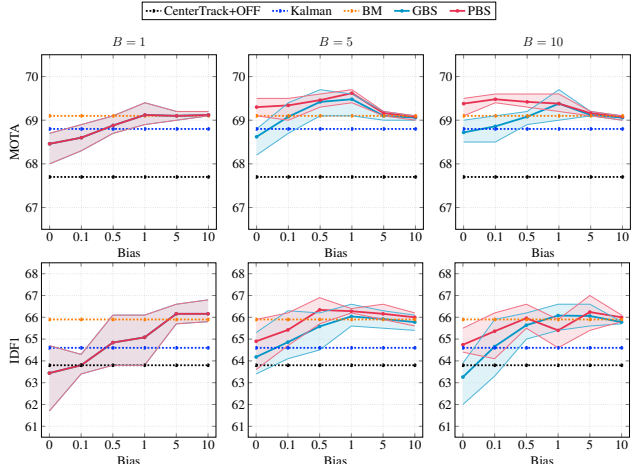
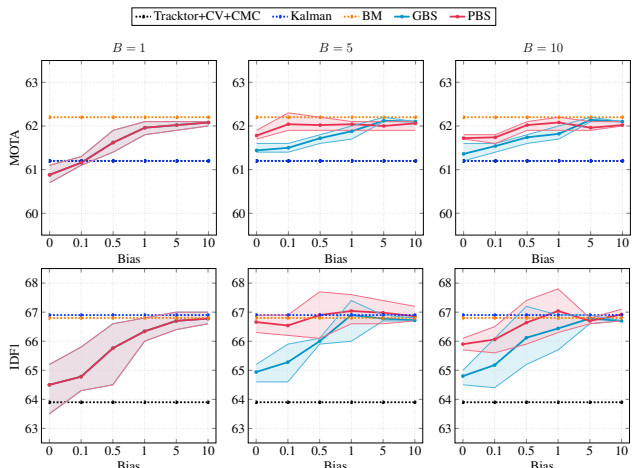
\begin{figure}
    \begin{minipage}[b]{\linewidth}
        \centering
        \pgfplotsset{compat = 1.3}
\pgfplotstableread{plots/bw_1/tracktor_pbs_OVERALL.tex}\varpbsone
\pgfplotstableread{plots/bw_3/tracktor_pbs_OVERALL.tex}\varpbsthree
\pgfplotstableread{plots/bw_5/tracktor_pbs_OVERALL.tex}\varpbsfive
\pgfplotstableread{plots/bw_10/tracktor_pbs_OVERALL.tex}\varpbsten
\pgfplotstableread{plots/bw_1/tracktor_gbs_OVERALL.tex}\vargbsone
\pgfplotstableread{plots/bw_3/tracktor_gbs_OVERALL.tex}\vargbsthree
\pgfplotstableread{plots/bw_5/tracktor_gbs_OVERALL.tex}\vargbsfive
\pgfplotstableread{plots/bw_10/tracktor_gbs_OVERALL.tex}\vargbsten

\begin{center}
\begin{minipage}{\textwidth}
\resizebox{0.78\linewidth}{!}{\begin{tikzpicture}[/pgfplots/width=\linewidth, /pgfplots/height=\linewidth, /pgfplots/legend pos=south east]
\begin{axis}[
    	ymin=59.5, ymax=63.5,
        xmin=0, xmax=5,
        ylabel={\Large MOTA},
        title={\Large $B=1$},
        xlabel=Bias,
		font=\Large,
        grid=both,
		grid style=dotted,
        xmode=linear,
        ytick={59, ..., 63},
		yticklabels={59, ..., 63},
        xtick={0, 1, 2, 3, 4, 5},
	    xticklabels={0, 0.1, 0.5, 1, 5, 10},
        enlarge x limits=0.05,
        legend style={at={(2.6,1.2)}, /tikz/every even column/.append style={column sep=3.5mm}},
        legend columns = -1,
        ]
        \addlegendentry{Tracktor+CV+CMC}
		\addplot[color=c_tracktor,mark=*, mark size=1, line width=2, dotted]
		coordinates{(0,61.2)(1,61.2)(2,61.2)(3,61.2)(4,61.2)(5,61.2)};
        \addlegendentry{Kalman}
		\addplot[color=c_kalman,mark=*, mark size=1, line width=2, dotted]
		coordinates{(0,61.2)(1,61.2)(2,61.2)(3,61.2)(4,61.2)(5,61.2)};
		\addlegendentry{BM}
		\addplot[color=c_best_mean,mark=*, mark size=1, line width=2, dotted]
		coordinates{(0,62.2)(1,62.2)(2,62.2)(3,62.2)(4,62.2)(5,62.2)};
		
		\addlegendentry{GBS}
		\addplot[color=c_vgbs,mark=*, mark size=1, line width=2] table[x=num,y=MOTA_mean] {\vargbsone};
		\addlegendentry{PBS}
		\addplot[color=c_vpbs,mark=*, mark size=1, line width=2] table[x=num,y=MOTA_mean] {\varpbsone};
		
		\addplot[name path=us_top,color=c_vgbs!70] table[x=num,y=MOTA_max] {\vargbsone};
		\addplot[name path=us_down,color=c_vgbs!70] table[x=num,y=MOTA_min] {\vargbsone};
		\addplot[c_vgbs!25,fill opacity=0.5] fill between[of=us_top and us_down];
		\addplot[name path=us_top,color=c_vpbs!70] table[x=num,y=MOTA_max] {\varpbsone};
		\addplot[name path=us_down,color=c_vpbs!70] table[x=num,y=MOTA_min] {\varpbsone};
		\addplot[c_vpbs!25,fill opacity=0.5] fill between[of=us_top and us_down];

    \end{axis}
    
\begin{axis}[
    	ymin=59.5, ymax=63.5,
        xmin=0, xmax=5,
        title={\Large $B=5$},
        xlabel=Bias,
		font=\Large,
        grid=both,
		grid style=dotted,
        xmode=linear,
        ytick={59, ..., 63},
		yticklabels={59, ..., 63},
        xtick={0, 1, 2, 3, 4, 5},
	    xticklabels={0, 0.1, 0.5, 1, 5, 10},
        enlarge x limits=0.05,
        legend style={ at={(1,0)}},
        xshift=8cm
        ]

		\addplot[color=c_vgbs,mark=*, mark size=1, line width=2] table[x=num,y=MOTA_mean] {\vargbsfive};
		\addplot[color=c_vpbs,mark=*, mark size=1, line width=2] table[x=num,y=MOTA_mean] {\varpbsfive};

		\addplot[color=c_tracktor,mark=*, mark size=1, line width=2, dotted]
		coordinates{(0,61.2)(1,61.2)(2,61.2)(3,61.2)(4,61.2)(5,61.2)};
		\addplot[color=c_kalman,mark=*, mark size=1, line width=2, dotted]
		coordinates{(0,61.2)(1,61.2)(2,61.2)(3,61.2)(4,61.2)(5,61.2)};
		\addplot[color=c_best_mean,mark=*, mark size=1, line width=2, dotted]
		coordinates{(0,62.2)(1,62.2)(2,62.2)(3,62.2)(4,62.2)(5,62.2)};
		
		\addplot[name path=us_top,color=c_vgbs!70] table[x=num,y=MOTA_max] {\vargbsfive};
		\addplot[name path=us_down,color=c_vgbs!70] table[x=num,y=MOTA_min] {\vargbsfive};
		\addplot[c_vgbs!25,fill opacity=0.5] fill between[of=us_top and us_down];
		\addplot[name path=us_top,color=c_vpbs!70] table[x=num,y=MOTA_max] {\varpbsfive};
		\addplot[name path=us_down,color=c_vpbs!70] table[x=num,y=MOTA_min] {\varpbsfive};
		\addplot[c_vpbs!25,fill opacity=0.5] fill between[of=us_top and us_down];

    \end{axis}

\begin{axis}[
    	ymin=59.5, ymax=63.5,
        xmin=0, xmax=5,
        title={\Large $B=10$},
        xlabel=Bias,
		font=\Large,
        grid=both,
		grid style=dotted,
        xmode=linear,
        ytick={59, ..., 63},
		yticklabels={59, ..., 63},
        xtick={0, 1, 2, 3, 4, 5},
	    xticklabels={0, 0.1, 0.5, 1, 5, 10},
        enlarge x limits=0.05,
        legend style={ at={(1,0)}},
        xshift=16cm
        ]

		\addplot[color=c_vgbs,mark=*, mark size=1, line width=2] table[x=num,y=MOTA_mean] {\vargbsten};
		\addplot[color=c_vpbs,mark=*, mark size=1, line width=2] table[x=num,y=MOTA_mean] {\varpbsten};
        
		\addplot[color=c_tracktor,mark=*, mark size=1, line width=2, dotted]
		coordinates{(0,61.2)(1,61.2)(2,61.2)(3,61.2)(4,61.2)(5,61.2)};
		\addplot[color=c_kalman,mark=*, mark size=1, line width=2, dotted]
		coordinates{(0,61.2)(1,61.2)(2,61.2)(3,61.2)(4,61.2)(5,61.2)};
		\addplot[color=c_best_mean,mark=*, mark size=1, line width=2, dotted]
		coordinates{(0,62.2)(1,62.2)(2,62.2)(3,62.2)(4,62.2)(5,62.2)};
		
		\addplot[name path=us_top,color=c_vgbs!70] table[x=num,y=MOTA_max] {\vargbsten};
		\addplot[name path=us_down,color=c_vgbs!70] table[x=num,y=MOTA_min] {\vargbsten};
		\addplot[c_vgbs!25,fill opacity=0.5] fill between[of=us_top and us_down];
		\addplot[name path=us_top,color=c_vpbs!70] table[x=num,y=MOTA_max] {\varpbsten};
		\addplot[name path=us_down,color=c_vpbs!70] table[x=num,y=MOTA_min] {\varpbsten};
		\addplot[c_vpbs!25,fill opacity=0.5] fill between[of=us_top and us_down];

    \end{axis}

\end{tikzpicture}}
\end{minipage}
\end{center}
    \end{minipage}
    \begin{minipage}[b]{\linewidth}
        \centering
        \pgfplotsset{compat = 1.3}
\pgfplotstableread{plots/bw_1/tracktor_pbs_OVERALL.tex}\varpbsone
\pgfplotstableread{plots/bw_3/tracktor_pbs_OVERALL.tex}\varpbsthree
\pgfplotstableread{plots/bw_5/tracktor_pbs_OVERALL.tex}\varpbsfive
\pgfplotstableread{plots/bw_10/tracktor_pbs_OVERALL.tex}\varpbsten
\pgfplotstableread{plots/bw_1/tracktor_gbs_OVERALL.tex}\vargbsone
\pgfplotstableread{plots/bw_3/tracktor_gbs_OVERALL.tex}\vargbsthree
\pgfplotstableread{plots/bw_5/tracktor_gbs_OVERALL.tex}\vargbsfive
\pgfplotstableread{plots/bw_10/tracktor_gbs_OVERALL.tex}\vargbsten

\begin{center}
\begin{minipage}{\textwidth}
\resizebox{0.336\textwidth}{!}{\begin{tikzpicture}[/pgfplots/width=\textwidth, /pgfplots/height=\textwidth, /pgfplots/legend pos=south east]
\begin{axis}[
    	ymin=62.5, ymax=68.5,
        xmin=0, xmax=5,
        ylabel={\Large IDF1},
        xlabel=Bias,
		font=\Large,
        grid=both,
		grid style=dotted,
        xmode=linear,
        ytick={63, ..., 68},
		yticklabels={63, ..., 68},
        xtick={0, 1, 2, 3, 4, 5},
	    xticklabels={0, 0.1, 0.5, 1, 5, 10},
        enlarge x limits=0.05,
        legend style={ at={(1,0)},
        draw=dashed},
        ]
		\addplot[color=c_vgbs,mark=*, mark size=1, line width=2] table[x=num,y=IDF1_mean] {\vargbsone};
		\addplot[color=c_vpbs,mark=*, mark size=1, line width=2] table[x=num,y=IDF1_mean] {\varpbsone};

		\addplot[color=c_tracktor,mark=*, mark size=1, line width=2, dotted]
		coordinates{(0,63.9)(1,63.9)(2,63.9)(3,63.9)(4,63.9)(5,63.9)};
		\addplot[color=c_kalman,mark=*, mark size=1, line width=2, dotted]
		coordinates{(0,66.9)(1,66.9)(2,66.9)(3,66.9)(4,66.9)(5,66.9)};
		\addplot[color=c_best_mean,mark=*, mark size=1, line width=2, dotted]
		coordinates{(0,66.8)(1,66.8)(2,66.8)(3,66.8)(4,66.8)(5,66.8)};
		
		\addplot[name path=us_top,color=c_vgbs!70] table[x=num,y=IDF1_max] {\vargbsone};
		\addplot[name path=us_down,color=c_vgbs!70] table[x=num,y=IDF1_min] {\vargbsone};
		\addplot[c_vgbs!25,fill opacity=0.5] fill between[of=us_top and us_down];
		\addplot[name path=us_top,color=c_vpbs!70] table[x=num,y=IDF1_max] {\varpbsone};
		\addplot[name path=us_down,color=c_vpbs!70] table[x=num,y=IDF1_min] {\varpbsone};
		\addplot[c_vpbs!25,fill opacity=0.5] fill between[of=us_top and us_down];

    \end{axis}
    
\begin{axis}[
    	ymin=62.5, ymax=68.5,
        xmin=0, xmax=5,
        xlabel=Bias,
		font=\Large,
        grid=both,
		grid style=dotted,
        xmode=linear,
        ytick={63, ..., 68},
		yticklabels={63, ..., 68},
        xtick={0, 1, 2, 3, 4, 5},
	    xticklabels={0, 0.1, 0.5, 1, 5, 10},
        enlarge x limits=0.05,
        legend style={ at={(1,0)},
        draw=dashed},
        xshift=8cm
        ]

		\addplot[color=c_vgbs,mark=*, mark size=1, line width=2] table[x=num,y=IDF1_mean] {\vargbsfive};
		\addplot[color=c_vpbs,mark=*, mark size=1, line width=2] table[x=num,y=IDF1_mean] {\varpbsfive};

		\addplot[color=c_tracktor,mark=*, mark size=1, line width=2, dotted]
		coordinates{(0,63.9)(1,63.9)(2,63.9)(3,63.9)(4,63.9)(5,63.9)};
		\addplot[color=c_kalman,mark=*, mark size=1, line width=2, dotted]
		coordinates{(0,66.9)(1,66.9)(2,66.9)(3,66.9)(4,66.9)(5,66.9)};
		\addplot[color=c_best_mean,mark=*, mark size=1, line width=2, dotted]
		coordinates{(0,66.8)(1,66.8)(2,66.8)(3,66.8)(4,66.8)(5,66.8)};
		
		\addplot[name path=us_top,color=c_vgbs!70] table[x=num,y=IDF1_max] {\vargbsfive};
		\addplot[name path=us_down,color=c_vgbs!70] table[x=num,y=IDF1_min] {\vargbsfive};
		\addplot[c_vgbs!25,fill opacity=0.5] fill between[of=us_top and us_down];
		\addplot[name path=us_top,color=c_vpbs!70] table[x=num,y=IDF1_max] {\varpbsfive};
		\addplot[name path=us_down,color=c_vpbs!70] table[x=num,y=IDF1_min] {\varpbsfive};
		\addplot[c_vpbs!25,fill opacity=0.5] fill between[of=us_top and us_down];

    \end{axis}

\begin{axis}[
    	ymin=62.5, ymax=68.5,
        xmin=0, xmax=5,
        xlabel=Bias,
		font=\Large,
        grid=both,
		grid style=dotted,
        xmode=linear,
        ytick={63, ..., 68},
		yticklabels={63, ..., 68},
        xtick={0, 1, 2, 3, 4, 5},
	    xticklabels={0, 0.1, 0.5, 1, 5, 10},
        enlarge x limits=0.05,
        legend style={ at={(1,0)},
        draw=dashed},
        xshift=16cm
        ]

		\addplot[color=c_vgbs,mark=*, mark size=1, line width=2] table[x=num,y=IDF1_mean] {\vargbsten};
		\addplot[color=c_vpbs,mark=*, mark size=1, line width=2] table[x=num,y=IDF1_mean] {\varpbsten};

		\addplot[color=c_tracktor,mark=*, mark size=1, line width=2, dotted]
		coordinates{(0,63.9)(1,63.9)(2,63.9)(3,63.9)(4,63.9)(5,63.9)};
		\addplot[color=c_kalman,mark=*, mark size=1, line width=2, dotted]
		coordinates{(0,66.9)(1,66.9)(2,66.9)(3,66.9)(4,66.9)(5,66.9)};
		\addplot[color=c_best_mean,mark=*, mark size=1, line width=2, dotted]
		coordinates{(0,66.8)(1,66.8)(2,66.8)(3,66.8)(4,66.8)(5,66.8)};
		
		\addplot[name path=us_top,color=c_vgbs!70] table[x=num,y=IDF1_max] {\vargbsten};
		\addplot[name path=us_down,color=c_vgbs!70] table[x=num,y=IDF1_min] {\vargbsten};
		\addplot[c_vgbs!25,fill opacity=0.5] fill between[of=us_top and us_down];
		\addplot[name path=us_top,color=c_vpbs!70] table[x=num,y=IDF1_max] {\varpbsten};
		\addplot[name path=us_down,color=c_vpbs!70] table[x=num,y=IDF1_min] {\varpbsten};
		\addplot[c_vpbs!25,fill opacity=0.5] fill between[of=us_top and us_down];

    \end{axis}

\end{tikzpicture}}
\end{minipage}
\end{center}
    \end{minipage}
    \caption{Experiments over TrajE parameters $bias$ and beam width $B$ for Tracktor using TrajE or Kalman with occlusion reconstruction. The y-axis correspond to the metric score, the x-axis to the $bias$, and the columns to the different $B$ values. 
    Solid lines correspond to the average value over five runs, and the transparent region limits correspond to the maximum and minimum values.}
    \label{plot:tracktor_var}
    \vspace{-3mm}
\end{figure}


\noindent \textbf{Comparison with state of the art.} In Table \ref{tab:mot17} we compare the two trackers using TrajE with the PBS setting that lead to the best results on the training set ($bias=1$, $B=5$), with and without occlusion reconstruction, with respect to the state of the art of multiple object trackers in the test set of the MOTChallenge dataset.
By using TrajE to predict trajectories, their performance in both MOTA and IDF1 scores is boosted by a considerable margin, and set a new state of the art results in the case of CenterTrack + TrajE, with an increase of $5.9$, and $6.3$ points in the MOTA score without and with occlusion reconstruction respectively. 

\begin{table}[t!]
\center
\tabcolsep=0.07cm

    \resizebox{1.0\linewidth}{!}{
    \begin{tabular}{l c c c c c c }
     \toprule
     \multicolumn{7}{c}{ MOT17~\cite{MOT16} \vspace{1mm}  } \\
     \midrule
     Method & MOTA $\uparrow$ & IDF1 $\uparrow$ & MT \% $\uparrow$ & ML \% $\downarrow$ & FP $\downarrow$ & FN $\downarrow$ \\ [0.5ex] 
     \midrule
     \multicolumn{7}{l}{Online, TrajE without occlusion reconstruction} \\
     \midrule
     \textbf{CenterTrack+TrajE} &  \color{red}\textbf{67.4}(+5.9) & \color{red}\textbf{61.2}(+1.6) & \color{red}\textbf{34.8} &  \color{red}\textbf{24.9} & 18652 & \color{red}\textbf{161347}  \\
     
     CenterTrack \cite{zhou2020tracking} & \color{blue}{61.5} & \color{blue}59.6 & \color{blue}26.4 &  \color{blue}{31.9} & 14076 & \color{blue}{200672}\\
     GSM \cite{ijcai2020-74} & 56.4 & 57.8 & 22.2 & 34.5 & 14379 & 230174 \\
     \textbf{Tracktor(v2)+TrajE} & 56.3(+0.0) & 57.8(+2.7) & 21.4 & 35.8 & \color{blue}{10068} & 233885  \\
     Tracktor(v2)\cite{bergmann2019tracking} & 56.3 & 55.1 & 21.1 & 35.3 & \color{red}\textbf{8866} & 235449 \\
     FAMNet \cite{chu2019famnet} & 52.0 & 48.7 & 19.1 & 33.4 & 14138 & 253616  \\
     STRN \cite{xu2019spatial} & 50.9 & 56.0 & 18.9 & 33.8 & 25295 & 249365  \\
     
     \midrule
     \multicolumn{7}{l}{Offline, TrajE with occlusion reconstruction} \\
     \midrule
     \textbf{CenterTrack+TrajE} &  \color{red}\textbf{67.8}(+6.3) & 61.4(+1.8) & \color{red}\textbf{36.0} &  \color{red}\textbf{24.5} & 20982 & \color{red}\textbf{157468}  \\
     
     Lif\_T \cite{hornakova2020lifted}  & \color{blue}60.5 & \color{red}\textbf{65.6} & 27.0 & 33.6 & 14966  & \color{blue}206619 \\
     MPNTrack \cite{braso2020learning} & 58.8 & {61.7} & \color{blue}{28.8} & \color{blue}33.5 & 17413 & 213594 \\
     
     \textbf{Tracktor(v2)+TrajE} & 56.6(+0.3) & 58.2(+3.1) & 21.9 & 35.7 & \color{red}\textbf{10119} & 231091  \\
     
     
     TT17   \cite{zhang2020long}  & 54.9 & \color{blue}63.1 & 24.4 & 38.1 & 20236 & 233295  \\
     TPM    \cite{peng2020tpm}  & 54.2 & 52.6 & 22.8 & 37.5 & \color{blue}13739 & 242730  \\
     
     JBNOT \cite{henschel2019multiple} & 52.6 & 50.8 & 19.7 & 35.8 & 31572 & 232659  \\
        
     ETC \cite{wang2019exploit} & 51.9 & 58.1 & 23.1 & 35.5 & 36164 & 232783  \\   
     eHAF\cite{sheng2018heterogeneous} & 51.8 & 54.7 & 23.4 & 37.9 & 33212 & 236772 \\

     \bottomrule
    \end{tabular}}
\vspace{0.05cm}
\caption{Comparison integrating \textit{TrajE} to CenterTrack and Tracktor against the state of the art methods on the test set of MOTChallenge 17 dataset using public detections. In bold tracker + our method (TrajE). In red the best result, in blue the second best. The tracking baselines are CenterTrack+OFF, Tracktor+CV+CMC.}
\label{tab:mot17}
\vspace{-1mm}
\end{table}
\begin{table}[t!]
\center
\tabcolsep=0.07cm

    \resizebox{1.0\linewidth}{!}{
    \begin{tabular}{l c c c c c c}
     \toprule
     \multicolumn{7}{c}{UA-DETRAC~\cite{CVIU_UA-DETRAC} \vspace{1mm}} \\
     \midrule
     Method & MOTA $\uparrow$ & IDF1 $\uparrow$ & MT \% $\uparrow$ & ML \% $\downarrow$ & FP $\downarrow$ & FN $\downarrow$  \\ [0.5ex] 
     \midrule
    \midrule
    \multicolumn{7}{l}{Online, TrajE without occlusion reconstruction} \\
    \midrule
    \textbf{CenterTrack+TrajE} & \color{red}\textbf{69.9}(+0.1) & \color{red}\textbf{80.1}(+1.6) & \color{blue}{70.0} & 7.9 & 41664 & \color{red}\textbf{146722}  \\
    CenterTrack \cite{zhou2020tracking} & \color{blue}{69.8} & \color{blue}{78.4} & {69.9} & 7.9 & 40994 & \color{blue}{147244} \\
     
    POI \cite{yu2016poi} & 69.2 & - & 67.4 & 5.2 & - & -\\
    WD \cite{zeng2020robust} & 68.5 & - & \color{red}\textbf{70.2} & \color{red}\textbf{3.1} & - & -\\
    \textbf{Tracktor(v2)+TrajE} & 67.7(+0.1) & 75.5(+1.3) & 60.4 & 10.0 & \color{red}\textbf{23273} & 194040\\
    Tracktor(v2) \cite{bergmann2019tracking} & 67.6 & 74.2 & 60.3 & 10.1 & \color{blue}{23699} & 194234\\
    
    DeepSORT \cite{wojke2017simple} & 65.4 & - & 65.9 & 4.7 & - & -\\
    RMOT \cite{yoon2015bayesian} & 62.6 & - & 42.1 & 6.5 & - & -\\
    IHTLS \cite{dicle2013way} & 62.6 & - & 63.4 & \color{blue}{3.8} &  - & - \\
    \midrule
    \multicolumn{7}{l}{Offline, TrajE with occlusion reconstruction} \\
    \midrule
    \textbf{CenterTrack+TrajE} & \color{red}\textbf{70.3}(+0.5) & \color{red}\textbf{80.4}(+2.0) & \color{red}\textbf{70.3} & 7.9 & 42277 & \color{red}\textbf{143938}  \\
    \textbf{Tracktor(v2)+TrajE} & 68.9+(1.3) & 75.7(+1.5) & 63.3 & 9.0 & {30135} & 179468 \\
    
    \bottomrule
    \end{tabular}}
\vspace{0.05cm}
\caption{Comparison integrating \textit{TrajE} to CenterTrack and Tracktor with state-of-the art methods on the test set of UA-DETRAC dataset using CompACT \cite{cai2015learning} public detections. In bold tracker + our method (TrajE). In red the best result, in blue the second best. Results from \cite{zeng2020robust}.}
\vspace{-5mm}
\label{tab:ua-detrac}

\end{table}



To see if TrajE generalizes, we tested both trackers + TrajE trained with MOTChallenge data against the test set of the UA-DETRAC dataset. In Table \ref{tab:ua-detrac}, we compare both CenterTrack and Tracktor with and without TrajE in the PBS configuration without and with occlusion reconstruction, with respect to other state of the art results on the UA-DETRAC dataset. In both cases, using TrajE for trajectory estimation boosts the performance of the trackers. 
It is important to state that UA-DETRAC dataset does not contain fully annotated frames (i.e only some of the cars inside the scene are annotated), and some of the non-annotated cars are inside "ignore zones", where the detections should be disregarded. We modified the MOT evaluation tool to take into account such zones by ignoring the detections whose bounding box centroid is inside the "ignore zones".

\vspace{-1mm}
\section{Conclusions}
\label{sec:conclusions}
\vspace{-2mm}

We have introduced \textit{TrajE}, a lightweight trajectory estimator based on mixture density networks and beam search, capable of significantly increasing the performance of existing multiple object trackers. Also, with the same estimated trajectory, we propose to do a track reconstruction when the object is lost due occlusions.
Our experiments adding our trajectory estimator to \textit{CenterTrack}, and \textit{Tracktor} provide very interesting insights on how the trajectory estimation can help in the tracking, while establishing a new state of the art in the MOTChallenge and UA-DETRAC datasets.

{\small
\bibliographystyle{ieee_fullname}
\bibliography{bibliography}
}

\end{document}